%% file: main.tex
\documentclass[11pt]{article}

\usepackage{acl}

\usepackage{amsmath}
\usepackage{amssymb}

\usepackage{booktabs}
\usepackage{graphicx}
\usepackage{multirow}
\usepackage{array}

\usepackage{natbib}

\usepackage{tikz}
\usetikzlibrary{positioning, arrows.meta}
\usepackage{pgfplots}
\pgfplotsset{compat=1.18}

\newcommand{\engrama}{\textsc{Engrama}}
\newcommand{\engramabench}{\textsc{EngramaBench}}
\newcommand{\memzero}{\textsc{Mem0}}
\newcommand{\singlespace}{\texttt{single\allowbreak\_space}}
\newcommand{\crossspace}{\texttt{cross\allowbreak\_space}}
\newcommand{\temporalcrossspace}{\texttt{temporal\allowbreak\_cross\allowbreak\_space}}
\newcommand{\adversarial}{\texttt{adversarial}}
\newcommand{\emergentinsight}{\texttt{emergent\allowbreak\_insight}}

\newcommand{\repourl}{https://github.com/julianacunadc/engramabench}
\newcommand{\repotext}{Publicly available at}
\newcommand{\repostatement}{The benchmark data, scorer, and canonical results are publicly available.}
\hypersetup{
  pdftitle={EngramaBench: Evaluating Long-Term Conversational Memory with Structured Graph Retrieval},
  pdfauthor={Juli\'{a}n Acu\~{n}a}
}

\title{\engramabench: Evaluating Long-Term Conversational Memory\\with Structured Graph Retrieval}

\author{
  Juli\'{a}n Acu\~{n}a \\
  \texttt{julianacuna@gmail.com}
}

\date{}

\begin{document}
\maketitle

\begin{abstract}
Large language model assistants are increasingly expected to retain and reason over information accumulated across many sessions.
We introduce \engramabench, a benchmark for long-term conversational memory built around five personas, one hundred multi-session conversations, and one hundred fifty queries spanning factual recall, cross-space integration, temporal reasoning, adversarial abstention, and emergent synthesis.
We evaluate \engrama, a graph-structured memory system, against GPT-4o full-context prompting and \memzero, an open-source vector-retrieval memory system. All three use the same answering model (GPT-4o), isolating the effect of memory architecture.
GPT-4o full-context achieves the highest composite score (0.6186), while \engrama{} scores 0.5367 globally but is the only system to score higher than full-context prompting on \crossspace{} reasoning (0.6532 vs.\ 0.6291, $n{=}30$). \memzero{} is cheapest but substantially weaker (0.4809).
Ablations reveal that the components driving \engrama's cross-space advantage trade off against global composite score, exposing a systems-level tension between structured memory specialization and aggregate optimization.
\end{abstract}

\input{sections/01_intro}
\input{sections/02_related_work}
\input{sections/03_engramabench}
\input{sections/04_engrama}
\input{sections/05_setup}
\input{sections/06_results}
\input{sections/07_ablations}
\input{sections/08_conclusion}

\input{sections/09_limitations}

\input{sections/10_ethics}

\bibliography{refs}

\clearpage
\appendix
\input{sections/appendix}

\end{document}

%% file: sections/01_intro.tex
\section{Introduction}
\label{sec:intro}

Large language model assistants are increasingly expected to behave as if they remember. Users interact with them as persistent collaborators that accumulate context over weeks and months, requiring the system to retain, recover, update, and connect information from prior interactions. These are fundamentally long-term memory problems.

One natural approach is to provide the model with as much prior history as possible at answer time. Recent long-context models make this strategy viable, and it remains surprisingly strong. However, full-context prompting scales poorly in cost and treats memory as transcript inclusion rather than structured representation. This leaves open whether it is the right long-term solution.

This tension motivates two questions. First, \textit{what kinds of memory behavior actually matter} in long-term conversational systems? Second, \textit{when does structured memory provide an advantage over brute-force context inclusion or flat retrieval?} Existing benchmarks partially address these questions but often omit cross-space integration, temporal change, or abstention under multi-session interaction. Many memory systems are also difficult to analyze because they conflate memory construction, retrieval, and answer generation.

We present two contributions. First, we introduce \engramabench, a benchmark for long-term conversational memory built around five canonical personas, one hundred multi-session conversations, and one hundred fifty queries spanning factual recall (\singlespace), cross-space integration (\crossspace), temporal reasoning (\temporalcrossspace), adversarial abstention (\adversarial), and emergent synthesis (\emergentinsight). Second, we evaluate \engrama, a graph-structured memory system, against \textbf{GPT-4o full-context} and \textbf{\memzero} \citep{chhikara2025mem0}, an open-source vector-retrieval memory system.

The resulting picture is nuanced. Full-context prompting achieves the highest composite score (\textbf{0.6186}), but \engrama{} is the only system to score higher on \crossspace{} reasoning (\textbf{0.6532} vs.\ \textbf{0.6291}; $n{=}30$), while \memzero{} trails both (\textbf{0.4809}). Ablations reveal that the components driving \engrama's cross-space strength trade off against aggregate score, exposing a design tension that helps explain why memory architectures can appear promising in smaller evaluations but reveal costs under broader benchmarking.

Our contributions:
\begin{itemize}
    \item \engramabench, a benchmark with explicit coverage of five memory behavior families including cross-space reasoning as a first-class target.
    \item A controlled comparison showing that full-context prompting wins globally while \engrama{} scores highest on \crossspace{} at substantially lower query-time cost.
    \item Ablation analysis separating the mechanisms that improve cross-space behavior from those that maximize aggregate score.
\end{itemize}

\noindent \repostatement\footnote{\repotext: \url{\repourl}}

%% file: sections/02_related_work.tex
\section{Related Work}
\label{sec:related}

\paragraph{Memory-augmented assistants.}
One strong baseline for long-term conversational tasks is full-context prompting: providing the model with the entire prior history at answer time. This avoids deciding what to remember but scales poorly and offers little control over memory organization. This limitation motivates memory-augmented systems that explicitly index, compress, or retrieve information from prior interactions, including MemoryBank \citep{zhong2023memorybank}, MemGPT/Letta \citep{packer2023memgpt}, ReadAgent \citep{lee2024readagent}, PerLTQA \citep{du2024perltqa}, \memzero{} \citep{chhikara2025mem0}, and LongMemEval \citep{wu2025longmemeval}. Graph-structured retrieval systems such as GraphRAG \citep{edge2024graphrag} and HippoRAG \citep{gutierrez2024hipporag} explore how structured substrates can support multi-hop recall more effectively than flat retrieval. \engrama{} is closest in spirit to this line but differs in treating long-term conversational state as a graph-structured memory surface organized around entities, spaces, temporal traces, and cross-space associations.

\paragraph{Benchmarks for conversational memory.}
Early multi-session benchmarks such as MSC \citep{xu2022msc} emphasize coherence and persona consistency over limited sessions. LoCoMo \citep{maharana2024locomo} introduces very long multi-session conversations with temporal event graphs, showing that even strong systems struggle as histories grow. LongMemEval \citep{wu2025longmemeval} focuses on chat assistants and decomposes memory into indexing, retrieval, and reading stages. However, prior benchmarks often underemphasize: (i) explicit integration across semantically distinct parts of a user's life, (ii) the interaction between temporal change and cross-session reasoning, (iii) abstention under fabricated queries, and (iv) synthesis whose answers are not stated verbatim in any session.

\paragraph{Positioning.}
\engramabench{} complements this prior work by making \textbf{cross-space reasoning} a first-class benchmark target alongside adversarial abstention and emergent synthesis under controlled persona histories. It trades breadth for interpretability: by organizing each persona around recurring semantic spaces, it distinguishes memory behaviors that are often conflated under a single long-context score.

%% file: sections/03_engramabench.tex
\section{\engramabench}
\label{sec:engramabench}

We introduce \engramabench, a benchmark for evaluating long-term conversational memory under multi-session interaction. It tests not only isolated fact recovery but also cross-space integration, temporal reasoning, abstention under fabricated queries, and emergent synthesis from dispersed evidence.

\subsection{Problem Formulation}

Each benchmark instance is a tuple $(H, q, a, E)$: a chronologically ordered interaction history $H$, a query $q$ posed after that history, a reference answer $a$, and annotated evidence $E$. The full benchmark contains \textbf{5 personas}, \textbf{100 conversations}, and \textbf{150 queries}. Each persona contributes 20 conversations and 30 queries. Queries are posed only after the full history has been observed, so the task is delayed long-term recall rather than next-turn prediction.

\subsection{Dataset Construction}

The benchmark is organized around five canonical personas: \textbf{Priya Sharma} (PI/professor), \textbf{Carlos Mendoza} (startup founder), \textbf{Sam Torres} (artist/musician), \textbf{Diane Chen} (product lead), and \textbf{Kai Nakamura} (PhD student). Each persona is defined over five recurring semantic spaces that partition the major domains of their life. This space structure creates a memory landscape where some questions are resolvable within a single region while others require combining evidence across regions.

The personas span different professional domains, life stages, and discourse styles, reducing overfitting to a single archetype. For each persona, 20 timestamped conversations trace a medium-horizon narrative (late 2025 to early 2026). Conversations are synthetically authored through an LLM-assisted generation pipeline constrained by persona blueprints, cross-space threads, and temporal arcs, then checked for structural consistency.

Query generation follows a fixed per-persona distribution: \textbf{6} \singlespace, \textbf{6} \crossspace, \textbf{6} \temporalcrossspace, \textbf{4} \emergentinsight, and \textbf{8} \adversarial{} queries (30 total per persona).

\subsection{Query Taxonomy}

\engramabench{} evaluates five task families:

\begin{itemize}
    \item \textbf{\singlespace}: factual recall within a single semantic space.
    \item \textbf{\crossspace}: integration across two or more spaces, requiring linking information from distinct life domains.
    \item \textbf{\temporalcrossspace}: temporal reasoning over information distributed across spaces and time.
    \item \textbf{\adversarial}: fabricated or false-premise questions requiring abstention.
    \item \textbf{\emergentinsight}: synthesis questions whose answers require combining multiple pieces of evidence into a higher-level conclusion.
\end{itemize}

This taxonomy separates memory abilities that are often conflated. A system may excel at local recall yet fail at cross-space integration, or retrieve broadly but perform poorly on unsupported questions.

\subsection{Scoring}

Each query includes \texttt{evidence\_conversation\_ids} and \texttt{evidence\_message\_ids} for auditability. The scorer supports seven answer types: \texttt{entity}, \texttt{date}, \texttt{number}, \texttt{set}, \texttt{short\_span}, \texttt{abstain}, and \texttt{insight}. These are aggregated into the five task-family metrics. The composite score is:
\[
\begin{aligned}
S_{\text{composite}} ={}& 0.5 \cdot \tfrac{1}{3}(S_{\text{single}} + S_{\text{cross}} + S_{\text{temporal}}) \\
&+ 0.25 S_{\text{adversarial}} + 0.25 S_{\text{emergent}}
\end{aligned}
\]
This weighting gives equal total weight to factual memory, adversarial abstention, and synthesis. We note that \emergentinsight{} is currently scored with token-level F1 over a reference answer rather than a judge-based rubric, and should be interpreted more cautiously than the factual slices.

%% file: sections/04_engrama.tex
\section{\engrama}
\label{sec:engrama}

\engrama{} is a graph-structured long-term memory system for conversational retrieval and reasoning. Rather than treating memory as a flat store of extracted snippets, it organizes accumulated interaction traces into a structured representation preserving cross-space associations, temporal continuity, and entity-level coherence. We describe the system at a level sufficient to interpret the empirical results, omitting implementation-specific heuristics and scoring formulas for confidentiality.

\subsection{Architecture}

\engrama{} processes timestamped user-assistant conversations and maintains a graph-structured memory organized around four classes of information: \textbf{entities} (identity continuity across mentions), \textbf{semantic spaces} (distinct regions of a user's life), \textbf{temporal traces} (order and recency of events), and \textbf{associative links} (cross-space connections). Figure~\ref{fig:system_overview} illustrates the pipeline.

\input{figures/figure1_system_overview}

This design reflects a core observation in \engramabench: many difficult questions require recovering a \textit{configuration} of related memories rather than a single fact. A graph supports activation over relations and neighborhoods rather than only over isolated text chunks.

\subsection{Ingestion and Retrieval}

Each new conversation is processed into a structured memory update that identifies salient entities, relations, and contextual signals and attaches them to the existing graph. The representation retains \textbf{space information} and \textbf{temporal information} as first-class properties, enabling the system to distinguish a user's work commitments from family obligations while permitting cross-region interactions.

At query time, \engrama{} performs \textbf{memory activation} (identifying a relevant subregion) followed by \textbf{answer-time composition} (converting the retrieved subgraph into response-compatible context). Retrieval is guided by the entities, spaces, and temporal cues in the query and expands outward to collect jointly useful memory. The answering stage receives a focused, structured summary rather than the full graph.

\subsection{Configurable Design Levers}

We study two configurable levers.\footnote{An additional lever, L2, exists as an intermediate evidence-structuring layer but was disabled in all reported runs.}

\textbf{Lever 1 (L1)} combines \textbf{entity-first activation} with a \textbf{query planner}, biasing retrieval toward memory regions structurally aligned with the query's entities and semantic intent.

\textbf{Lever 3 (L3)} is a \textbf{typed answer layer} that strengthens the conversion of retrieved evidence into an explicit answer candidate when the task demands a specific form. It targets failures where the system retrieves relevant evidence but emits a vague response.

These levers separate two common failure sources: (1) retrieving the wrong memory region, and (2) failing to convert correct memory into a precise output. The ablation results in Section~\ref{sec:ablations} show that both help preserve cross-space performance but do not improve global composite in their current implementations.

%% file: figures/figure1_system_overview.tex
\begin{figure*}[t]
\centering
\begin{tikzpicture}[
    node distance=0.6cm,
    block/.style={
        rectangle, draw=black!60, fill=white,
        minimum height=2.6cm, minimum width=2.3cm,
        text width=2.0cm, align=center,
        font=\small, rounded corners=2pt,
        line width=0.5pt
    },
    centerblock/.style={
        rectangle, draw=black!70, fill=blue!4,
        minimum height=3.0cm, minimum width=3.0cm,
        text width=2.7cm, align=center,
        font=\small, rounded corners=2pt,
        line width=0.7pt
    },
    label/.style={
        font=\footnotesize\bfseries, text=black!80
    },
    sublabel/.style={
        font=\scriptsize, text=black!55
    },
    arrow/.style={
        ->, >=stealth, line width=0.8pt, draw=black!50
    },
    graphone/.style={
        circle, fill=blue!25, draw=blue!40,
        minimum size=5pt, inner sep=0pt
    },
    graphtwo/.style={
        circle, fill=blue!12, draw=blue!30,
        minimum size=4pt, inner sep=0pt
    }
]

\node[block] (conv) {
    {\scriptsize\texttt{Session 1}}\\[1pt]
    {\scriptsize\texttt{Session 2}}\\[1pt]
    {\scriptsize\texttt{...}}\\[1pt]
    {\scriptsize\texttt{Session $n$}}
};
\node[label, above=0.15cm of conv] {Conversations};
\node[sublabel, below=0.15cm of conv] {Multi-session history};

\node[block, right=0.7cm of conv] (ingest) {
    {\scriptsize Extract entities,}\\
    {\scriptsize relations, events}\\[4pt]
    {\scriptsize Attach to existing}\\
    {\scriptsize memory structure}
};
\node[label, above=0.15cm of ingest] {Ingestion};
\node[sublabel, below=0.15cm of ingest] {Consolidation};

\node[centerblock, right=0.7cm of ingest] (graph) {};
\node[label, above=0.15cm of graph] {Graph Memory};
\node[sublabel, below=0.15cm of graph] {Entities\enspace$\cdot$\enspace Spaces\enspace$\cdot$\enspace Temporal\enspace$\cdot$\enspace Links};

\node[graphone] (a1) at ([xshift=-0.55cm, yshift=0.45cm]graph.center) {};
\node[graphtwo] (a2) at ([xshift=-0.2cm, yshift=0.7cm]graph.center) {};
\node[graphtwo] (a3) at ([xshift=-0.85cm, yshift=0.2cm]graph.center) {};
\draw[blue!30, line width=0.4pt] (a1) -- (a2);
\draw[blue!30, line width=0.4pt] (a1) -- (a3);

\node[graphone] (b1) at ([xshift=0.5cm, yshift=-0.25cm]graph.center) {};
\node[graphtwo] (b2) at ([xshift=0.85cm, yshift=0.0cm]graph.center) {};
\node[graphtwo] (b3) at ([xshift=0.25cm, yshift=-0.5cm]graph.center) {};
\draw[blue!30, line width=0.4pt] (b1) -- (b2);
\draw[blue!30, line width=0.4pt] (b1) -- (b3);

\node[graphone] (c1) at ([xshift=0.55cm, yshift=0.55cm]graph.center) {};
\node[graphtwo] (c2) at ([xshift=0.9cm, yshift=0.35cm]graph.center) {};
\draw[blue!30, line width=0.4pt] (c1) -- (c2);

\draw[blue!20, line width=0.4pt, dashed] (a1) -- (b1);
\draw[blue!20, line width=0.4pt, dashed] (a2) -- (c1);
\draw[blue!20, line width=0.4pt, dashed] (b2) -- (c2);

\node[block, right=0.7cm of graph] (retrieval) {
    {\scriptsize Query-conditioned}\\
    {\scriptsize activation}\\[4pt]
    {\scriptsize Neighborhood}\\
    {\scriptsize retrieval}\\[4pt]
    {\scriptsize Structured summary}
};
\node[label, above=0.15cm of retrieval] {Retrieval};
\node[sublabel, below=0.15cm of retrieval] {Composition};

\node[block, right=0.7cm of retrieval, minimum width=2.0cm, text width=1.7cm] (answer) {
    {\scriptsize Answer grounded}\\
    {\scriptsize in retrieved}\\
    {\scriptsize memory region}
};
\node[label, above=0.15cm of answer] {Answer};

\draw[arrow] (conv.east) -- (ingest.west);
\draw[arrow] (ingest.east) -- (graph.west);
\draw[arrow] (graph.east) -- (retrieval.west);
\draw[arrow] (retrieval.east) -- (answer.west);

\node[sublabel] (query) at ([yshift=-1.0cm]retrieval.south) {\textit{Query}};
\draw[arrow, dashed] (query.north) -- (retrieval.south);

\end{tikzpicture}
\caption{High-level overview of \engrama. Conversations are incrementally consolidated into a graph-structured memory organized around entities, semantic spaces, temporal traces, and cross-space associative links (dashed edges). At query time, the system activates a relevant neighborhood rather than replaying the full transcript, converts it into a structured summary, and generates an answer grounded in that region. This graph-based retrieval path is consistent with \engrama's cross-space advantage over flat retrieval and full-context prompting.}
\label{fig:system_overview}
\end{figure*}
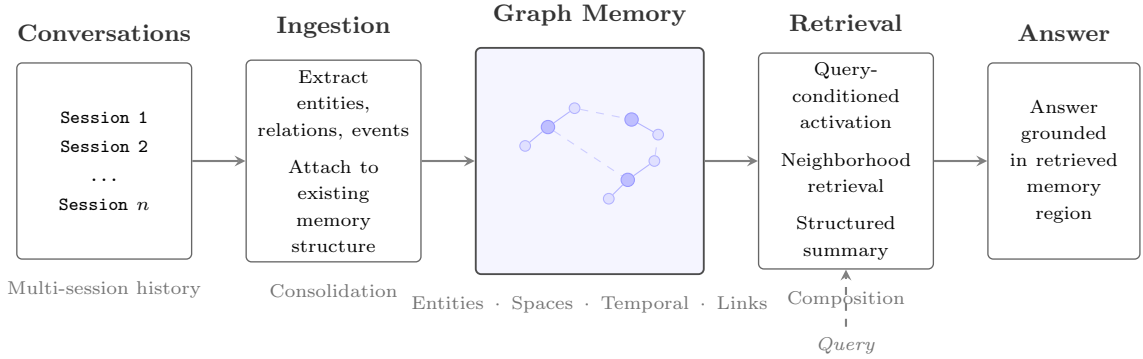

%% file: sections/05_setup.tex
\section{Experimental Setup}
\label{sec:setup}

All results use \texttt{EngramaBench full\_v1} (5 personas, 100 conversations, 150 queries) with the v1.4.0 scorer and unified entity registry. For \engrama, each query runs against a clean persona-specific canonical memory snapshot to avoid cross-query contamination.

\paragraph{Systems.}
All three systems use \texttt{gpt-4o-2024-08-06} at temperature 0.3 as the answering model, ensuring that benchmark differences reflect memory architecture rather than model quality.

\textbf{GPT-4o full-context}: all persona conversations are concatenated chronologically and injected into the prompt with no compression or retrieval.

\textbf{\memzero{}} \citep{chhikara2025mem0}: an open-source memory system using vector retrieval. We seed it from canonical histories, store extracted memories in ChromaDB, and retrieve top-$k{=}20$ memories at query time using \texttt{text-embedding-3-small}, matching its default configuration to evaluate an off-the-shelf vector memory baseline.

\textbf{\engrama}: the graph-structured memory system described in Section~\ref{sec:engrama}, evaluated in its benchmark configuration.

\paragraph{Ablations.}
We report three \engrama{} ablations: \texttt{no-L1} (disables entity-first activation + query planning), \texttt{no-L3} (disables the typed answer layer), and \texttt{no-L1-L3} (disables both).

\paragraph{Metrics.}
We report the five task-family metrics and the aggregate composite score defined in Section~\ref{sec:engramabench}. Per-persona composite scores are reported in Appendix~\ref{sec:app_persona}.

\paragraph{Cost.}
Query-time cost across all 150 queries: GPT-4o \textbf{\$3.33}, \engrama{} \textbf{\$0.67}, \memzero{} \textbf{\$0.36}. These figures represent online serving cost only; offline memory construction costs are excluded for all systems. See Appendix~\ref{sec:app_cost} for details.

%% file: sections/06_results.tex
\section{Results}
\label{sec:results}

\input{tables/table1_main_results}

\subsection{Overall Comparison}

Table~\ref{tab:main_results} and Figure~\ref{fig:main_results} show that \textbf{GPT-4o full-context} achieves the highest global score (\textbf{0.6186}). \textbf{\engrama{} full} scores \textbf{0.5367}, while \textbf{\memzero} reaches \textbf{0.4809}. Because all three systems use the same answering model, these differences reflect memory and retrieval architecture rather than model quality.

GPT-4o full-context is strongest on \singlespace{} (0.7339), \temporalcrossspace{} (0.3902), and \emergentinsight{} (0.3305). \memzero{} is weakest overall, especially on \singlespace{} (0.2848) and \temporalcrossspace{} (0.2356), though it achieves perfect adversarial abstention (1.0000). The performance gap between \memzero{} and the other systems is notable: even with the same answering model, flat vector retrieval over extracted memories is insufficient. This suggests that retrieval alone is not the bottleneck; memory construction and representation matter as well.

\input{figures/figure2_main_results}

\subsection{Cross-Space Advantage}

The clearest signature of \engrama's design appears in \crossspace{} performance. \textbf{\engrama{} full} achieves \textbf{0.6532}, compared to \textbf{0.6291} for GPT-4o and \textbf{0.5266} for \memzero{} ($n{=}30$ queries). While the sample size limits definitive conclusions, this is the only task family where structured memory scores higher than brute-force context inclusion.

This result matters because \crossspace{} is the most diagnostic family for structured long-term memory. A strong \singlespace{} score can come from local recall or brute-force transcript access; a strong \crossspace{} score requires something more compositional---identifying which parts of memory belong together even when evidence is not co-located. \engrama's graph structure preserves associations that flat retrieval or full-context prompting do not exploit as effectively.

The contrast with \memzero{} is especially informative: despite using the same answering model, \memzero{} trails \engrama{} by \textbf{12.7 points} on \crossspace{} (0.5266 vs.\ 0.6532), suggesting that the difference is between \textbf{structured retrieval and flat retrieval}, not merely retrieval versus no retrieval.

\subsection{Cost--Quality Tradeoff}

\engrama{} reaches about \textbf{86.8\%} of GPT-4o's composite at roughly \textbf{20\%} of the query-time cost, while scoring higher on the benchmark's \crossspace{} slice. \memzero{} is cheaper still but reaches only 77.7\% of GPT-4o's composite, falling behind \engrama{} on all factual families. \engrama{} therefore occupies the most compelling cost--quality position among the memory-augmented systems evaluated.

%% file: tables/table1_main_results.tex
\begin{table*}[t]
\centering
\small
\begin{tabular}{@{}lccccccc@{}}
\toprule
\textbf{System} & \texttt{single} & \texttt{cross} & \texttt{temporal} & \texttt{advers.} & \texttt{emergent} & \textbf{Composite} & \textbf{Cost} \\
\midrule
GPT-4o full-context    & \textbf{0.7339} & 0.6291          & \textbf{0.3902} & 0.9750          & \textbf{0.3305} & \textbf{0.6186} & \$3.33 \\
\engrama{} full        & 0.5997          & \textbf{0.6532} & 0.3230          & 0.8000          & 0.2963          & 0.5367          & \$0.67 \\
\memzero{}             & 0.2848          & 0.5266          & 0.2356          & \textbf{1.0000} & 0.2255          & 0.4809          & \$0.36$^{\dagger}$ \\
\bottomrule
\end{tabular}
\caption{Main results on \engramabench{} \texttt{full\_v1} (150 queries, 5 personas). All three systems use GPT-4o as the answering model; differences therefore reflect memory architecture, not model quality. GPT-4o full-context leads globally, but \engrama{} achieves the highest \texttt{cross} score---the only metric where structured memory scores higher than brute-force context inclusion. \memzero{} is cheapest but weakest across all factual families. Cost is total query-time cost across 150 queries. Ablated \engrama{} variants are reported separately in Table~\ref{tab:ablations}. Bold indicates best per column. ${}^{\dagger}$\memzero{} cost excludes offline memory extraction and embedding costs.}
\label{tab:main_results}
\end{table*}

%% file: figures/figure2_main_results.tex
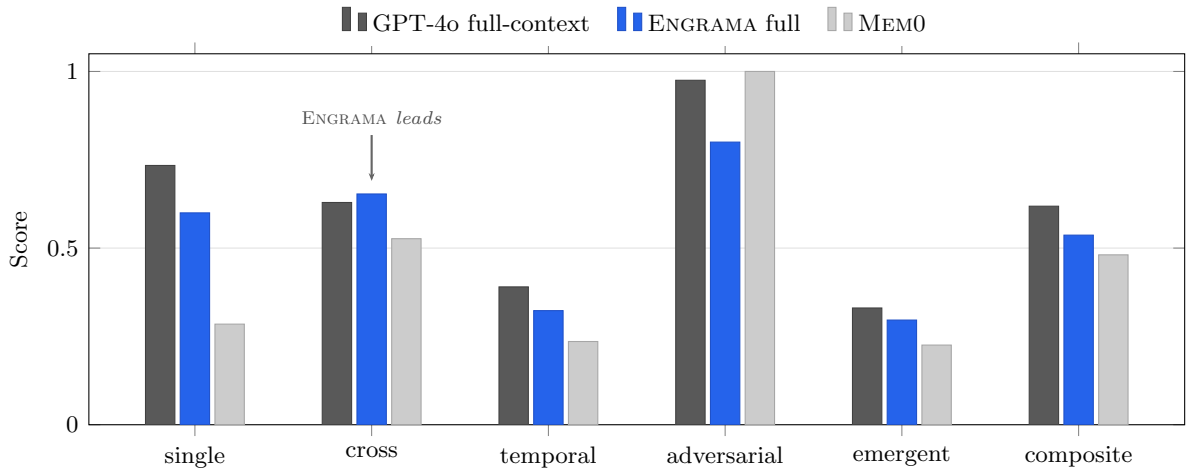
\begin{figure*}[t]
\centering
\begin{tikzpicture}
\pgfplotsset{
    compat=1.18,
    every axis/.style={
        width=\textwidth,
        height=6.5cm,
        ymin=0, ymax=1.05,
        ylabel={Score},
        ylabel style={font=\small},
        yticklabel style={font=\footnotesize},
        xticklabel style={font=\footnotesize},
        legend style={
            font=\footnotesize,
            at={(0.5,1.02)},
            anchor=south,
            legend columns=3,
            /tikz/every even column/.append style={column sep=8pt},
            draw=none,
        },
        ymajorgrids=true,
        grid style={line width=0.3pt, draw=gray!25},
        bar width=11pt,
        enlarge x limits=0.12,
        xtick=data,
        symbolic x coords={single, cross, temporal, adversarial, emergent, composite},
    }
}
\begin{axis}[ybar,
    legend entries={GPT-4o full-context, \engrama{} full, \memzero{}},
    ]
    \addplot[fill=black!65, draw=black!75] coordinates {
        (single, 0.7339)
        (cross, 0.6291)
        (temporal, 0.3902)
        (adversarial, 0.9750)
        (emergent, 0.3305)
        (composite, 0.6186)
    };
    \addplot[fill={rgb,255:red,37;green,99;blue,235}, draw={rgb,255:red,30;green,80;blue,200}] coordinates {
        (single, 0.5997)
        (cross, 0.6532)
        (temporal, 0.3230)
        (adversarial, 0.8000)
        (emergent, 0.2963)
        (composite, 0.5367)
    };
    \addplot[fill=black!20, draw=black!35] coordinates {
        (single, 0.2848)
        (cross, 0.5266)
        (temporal, 0.2356)
        (adversarial, 1.0000)
        (emergent, 0.2255)
        (composite, 0.4809)
    };

    \draw[-{Stealth[length=3pt]}, thick, black!60]
        (axis cs:cross, 0.82) -- (axis cs:cross, 0.69)
        node[pos=0, above, font=\scriptsize\itshape, text=black!70, align=center]
        {\engrama{} leads};
\end{axis}
\end{tikzpicture}
\caption{Per-metric comparison of the three primary systems on \engramabench{} (150 queries). GPT-4o full-context (dark) leads on most metrics and overall composite. \engrama{} (blue) is the only system to outperform full-context prompting, doing so on \texttt{cross}---the metric most diagnostic of structured long-term memory. \memzero{} (light) is weakest on factual families but achieves perfect adversarial abstention. Exact values in Table~\ref{tab:main_results}.}
\label{fig:main_results}
\end{figure*}

%% file: sections/07_ablations.tex
\section{Ablations}
\label{sec:ablations}

The main comparison establishes \textit{what} happens; the ablation study addresses \textit{why}. It reveals that the components driving the cross-space advantage are not the same ones that maximize composite score.

\input{tables/table3_ablations}

\subsection{Lever Effects}

Table~\ref{tab:ablations} shows the ablation results relative to \engrama{} full (composite = 0.5367, \crossspace{} = 0.6532). Removing \textbf{L1} raises composite to 0.5488 (+0.0121) and \singlespace{} to 0.6330 (+0.0333), but lowers \crossspace{} to 0.6368 ($-$0.0164). Removing \textbf{L3} produces a sharper version of the same pattern: composite rises to 0.5501 (the highest \engrama{} variant), but \crossspace{} drops nearly 5 points to 0.6033. Both levers therefore exchange global composite for better cross-space behavior.

This indicates that L1 and L3 are not generic quality multipliers but specialized components whose value is concentrated in the benchmark slice that most strongly rewards structured retrieval.

\subsection{Interaction Effect}

The most revealing result is \texttt{no-L1-L3}. If the levers imposed independent overhead, removing both should yield a larger composite gain than removing either alone. Instead, \texttt{no-L1-L3} returns to exactly 0.5367---matching the full system---while \crossspace{} drops further to 0.5986. This indicates a real \textbf{interaction effect}: L1 and L3 are each beneficial in isolation on global composite relative to the double-ablation baseline, but their gains do not compose when both are enabled. On \crossspace, however, the pair is complementary: the full system outperforms either lever alone and substantially exceeds the double-ablation baseline.

This interaction helps explain the gap between pilot and full benchmark results. In the smaller pilot, benefits appeared larger and more uniformly positive. The full benchmark, with more heterogeneous query families, exposes costs that become visible only at broader evaluation scale.

\subsection{Failure Modes}

Several recurring patterns emerge. \textbf{Temporal reasoning remains difficult for all systems}: even GPT-4o reaches only 0.3902. \engrama{} exhibits a tension between structured retrieval and typed answer precision---some failures are retrieval misses, others are answer-construction failures over partially correct memory states. \engrama's lower adversarial abstention (0.8000 vs.\ 0.9750/1.0000) reflects a structural property of graph-based retrieval: multi-hop activation can surface structurally plausible but factually irrelevant evidence for fabricated queries.

%% file: tables/table3_ablations.tex
\begin{table}[t]
\centering
\small
\begin{tabular}{@{}lccc@{}}
\toprule
\textbf{Variant} & $\Delta$\textbf{composite} & $\Delta$\textbf{single} & $\Delta$\textbf{cross} \\
\midrule
no-L1      & $+$0.0121 & $+$0.0333 & $-$0.0164 \\
no-L3      & $+$0.0134 & $+$0.0325 & $-$0.0499 \\
no-L1-L3   & $+$0.0000 & $+$0.0325 & $-$0.0546 \\
\bottomrule
\end{tabular}
\caption{Ablation deltas relative to \engrama{} full (composite = 0.5367). The pattern reveals a design tension: removing L1 or L3 individually improves global composite but degrades \crossspace, the metric most diagnostic of structured memory. Removing both eliminates the composite gain while further degrading cross-space, indicating that the levers interact sub-additively on composite but complementarily on cross-space.}
\label{tab:ablations}
\end{table}

%% file: sections/08_conclusion.tex
\section{Discussion and Conclusion}
\label{sec:conclusion}

The benchmark exposes a regime in which memory organization itself changes system behavior. Cross-space queries require recovering configurations of related facts never co-located in a single conversation, and here the graph representation provides a measurable edge. That this advantage coexists with a lower global composite is itself informative: structured memory is not yet a drop-in replacement for full-context prompting, but it is a sharper tool for specific reasoning demands that transcript inclusion handles only incidentally.

The ablation results sharpen this point. Current memory architectures face a coordination problem: mechanisms that organize memory for cross-space use do not automatically maximize aggregate utility. The next frontier is not simply more memory structure, but \textbf{more selective} structure---stronger when needed, quieter when not. The comparison with \memzero{} reinforces this: flat vector retrieval loses substantial ground on factual precision, temporal reasoning, and cross-space composition despite using the same answering model, suggesting that \textit{what is remembered} and \textit{how it is structured} are inseparable design questions.

\paragraph{Future work.}
Three directions follow. First, the benchmark should be extended with finer-grained evidence provenance and stronger evaluation for synthesis answers. Second, memory architectures like \engrama{} should be improved so that structural strengths translate more consistently into global gains. Third, additional baselines and naturalistic conversational settings are needed to characterize the boundary between brute-force context inclusion and genuinely effective long-term memory.

%% file: sections/09_limitations.tex

\section*{Limitations}
\label{sec:limitations}

\engramabench{} is built from synthetic but tightly constrained conversations rather than naturally occurring human-assistant logs. We regard this as an acceptable design choice for controlled benchmarking---canonical personas, evidence annotations, and known cross-space arcs would be difficult to obtain from organic logs---but the benchmark should not be interpreted as a direct measurement of production behavior.

The benchmark is reproducible and stable at the run-artifact level: it uses fixed conversation snapshots, frozen query files, explicit evidence annotations, versioned registries, and deterministic scoring outputs. However, it does not fully audit runtime evidence provenance at the exact \texttt{message\_id} or \texttt{chunk\_id} level for every evaluated system. Current results support strong claims about comparative answer quality, but more limited claims about exact evidence recovery dynamics.

Some ability regions remain underexplored. Temporal reasoning is difficult for all systems, including full-context prompting, suggesting that the current query formulation captures a genuinely hard problem rather than a weakness of any one architecture. The \emergentinsight{} scorer uses token-level F1 over a reference answer rather than a judge-based rubric, and should be interpreted more cautiously than the factual slices. Additionally, per-family sample sizes are modest ($n{=}30$ for each of the three factual families), which limits the statistical power of pairwise comparisons on individual task families; the cross-space advantage we observe is directionally consistent across personas but should be validated on larger query sets in future work.

There is a deliberate disclosure boundary in this paper. \engrama{} is a proprietary system, and we describe its architecture only at the level needed to make the empirical claims interpretable. We report structural design, benchmark behavior, and ablation surfaces, but omit implementation-specific retrieval heuristics, graph weighting formulas, internal prompts, and production orchestration details.

%% file: sections/10_ethics.tex

\section*{Ethical Considerations}
\label{sec:ethics}

Long-term memory systems for conversational assistants raise considerations around user privacy and data retention. By design, such systems accumulate and persist personal information across sessions, which creates potential risks if deployed without appropriate safeguards around data access, deletion, and consent.

This paper evaluates memory architectures using entirely synthetic personas and conversations. No real user data was collected, stored, or used at any stage of the benchmark construction or evaluation. The five canonical personas were designed to represent diverse professional and personal contexts without modeling any specific real individual.

The benchmark data, scorer, and canonical results will be released upon publication to support reproducibility and further research in this area.

%% file: sections/appendix.tex
\section{Per-Persona Analysis}
\label{sec:app_persona}

\input{tables/table2_per_persona}

Per-persona results (Table~\ref{tab:per_persona}) reveal that the systems differ not only in average performance but also in \textbf{stability across user archetypes}. GPT-4o full-context is the most stable, with per-persona composite scores ranging from 0.6037 (Kai) to 0.6405 (Sam), a spread of only 0.0368. \engrama{} is less stable but reasonably consistent (0.4928--0.5716, spread 0.0788). \memzero{} is the most variable (0.3687--0.5563, spread 0.1876).

This instability suggests that flat memory extraction and retrieval are much more sensitive to persona domain and temporal structure than either full-context prompting or graph-structured memory. \textbf{Kai} is consistently the hardest persona for memory baselines, likely because this persona contains substantial interaction between research, coursework, wellness, and career preparation. Diane is comparatively strong for both \engrama{} and \memzero, suggesting that some narratives compress more easily into stable memory representations.

\section{Cost Accounting Details}
\label{sec:app_cost}

Across all 150 benchmark queries, GPT-4o full-context costs approximately \$3.33, \engrama{} full approximately \$0.67, and \memzero{} approximately \$0.36 in query-time cost. These figures use frozen pricing tables in the evaluation runners and represent online serving cost under fixed benchmark conditions.

One caveat is important. The reported \memzero{} cost is query-time only and excludes offline memory seeding, extraction, and embedding generation. Likewise, \engrama{} figures do not amortize canonical snapshot construction. We interpret the cost numbers as serving-time comparisons between three answer-time regimes: brute-force full-context prompting, graph-structured memory retrieval, and vector retrieval over extracted memories.


%% file: tables/table2_per_persona.tex
\begin{table}[t]
\centering
\footnotesize
\resizebox{\columnwidth}{!}{%
\begin{tabular}{@{}lccc@{}}
\toprule
\textbf{Persona} & \textbf{\engrama} & \textbf{GPT-4o} & \textbf{$\Delta$} \\
\midrule
Priya (PI / prof.)             & 0.5716 & 0.6066 & $-$0.0350 \\
Carlos (founder)                & 0.5294 & 0.6397 & $-$0.1103 \\
Sam (artist / music.)           & 0.5569 & 0.6405 & $-$0.0836 \\
Diane (product lead)            & 0.5329 & 0.6052 & $-$0.0723 \\
Kai (PhD student)               & 0.4928 & 0.6037 & $-$0.1109 \\
\midrule
\textbf{Mean}                   & \textbf{0.5367} & \textbf{0.6191} & \textbf{$-$0.0824} \\
\bottomrule
\end{tabular}
}
\caption{Per-persona composite scores for \engrama{} full vs.\ GPT-4o full-context. Kai is consistently the hardest persona for \engrama; the gap is smallest for Priya. The GPT-4o mean here (0.6191) is a macro-average over persona-level composites; the global composite in Table~\ref{tab:main_results} (0.6186) is computed over all queries directly, so the two may differ slightly.}
\label{tab:per_persona}
\end{table}